%% file: main.tex
\documentclass[11pt]{article}
\usepackage{enumitem}

\usepackage[preprint]{acl}
\usepackage{times}
\usepackage{latexsym}
\usepackage[T1]{fontenc}
\usepackage[utf8]{inputenc}
\usepackage{microtype}
\usepackage{inconsolata}
\usepackage{amsthm}

\usepackage{graphicx}
\usepackage{algorithm}
\usepackage{algorithmic}
\usepackage{amsmath}
\usepackage{amssymb}
\usepackage{booktabs}
\usepackage{multirow}
\usepackage{fvextra}
\usepackage{tikz}
\usepackage[most]{tcolorbox}
\definecolor{cream}{RGB}{255,253,208}
\usepackage{listings}

\newtcblisting{promptbox}{
  enhanced,
  breakable,
  listing only,
  colback=gray!7,
  colframe=black!60,
  boxrule=0.6pt,
  arc=2mm,
  left=3mm,
  right=3mm,
  top=2mm,
  bottom=2mm,
  listing options={
    basicstyle=\ttfamily\footnotesize,
    breaklines=true,
    breakatwhitespace=true,
    columns=fullflexible,
    keepspaces=true,
    showstringspaces=false
  }
}

\tcbset{
  takeawaystyle/.style={
    colback=cream,               
    colframe=black!30,           
    boxrule=0.4pt,               
    arc=3pt,                     
    left=6pt, right=6pt, top=6pt, bottom=6pt, 
    fonttitle=\bfseries,         
    coltitle=black,              
    enhanced,                    
    drop shadow={shadow xshift=1pt, shadow yshift=-1pt, opacity=0.15}
  }
}

\title{ROAM: Robust Organization of Atomic Memories for Agents through Semantic Relations}
\newcommand{\cameraReadyAuthors}{%
  Jianjie Zheng$^{1*}$,
  Peng Lai$^{1}$\thanks{\ \ Equal Contribution.},
  Sijie Cheng$^{2,3}$,
  Jiehui Zhao$^{4}$, \
  Lei Yang$^{4}$, \
  Guanhua Chen$^{1}$\thanks{\ \ Corresponding Author.} \\
  $^1$Southern University of Science and Technology, $^2$Tsinghua University \\  
  $^3$RayNeo.AI,
  $^4$Deepexi Technology Co. Ltd. \\
}
\makeatletter
\ifacl@anonymize
  \author{Anonymous Submission}
\else
  \author{\cameraReadyAuthors}
\fi
\makeatother

\begin{document}
\maketitle

\begin{abstract}
Long-term language-model agents rely on external memory across interactions. Atomic memories are particularly useful: their fine-grained semantic boundaries enable precise retrieval and direct comparison between observations. Yet accumulating atoms inevitably become redundant, overlapping, or conflicting. Existing methods often ask an LLM manager to add, update, delete, or rewrite memories directly, coupling semantic interpretation, storage decisions, and content generation in one error-prone operation.
We introduce \emph{ROAM}, a relation-guided framework that uses atomicity for management while allowing richer answer-time representations. ROAM classifies incoming--stored atom pairs as independent, equivalent, directionally subsuming, or conflicting, then organizes observations into active Primary and supporting Evidence roles. Fusion subsequently combines complementary details and temporal changes into compact, potentially non-atomic views. Only Primary views are retrieved for answering, preventing redundant or outdated atoms from competing independently.
Across models and evaluation settings, ROAM improves answer accuracy by up to 29.8 percentage points. Ablations show complementary benefits from different relations and consistent gains from fusion beyond role organization. Mechanism analysis further finds 15.6-point higher answer-critical source recall and an 11.5-point lower confounder-token share. ROAM remains robust across manager scales.
\end{abstract}

\input{sections/01_introduction}
\input{sections/03_method}
\input{sections/04_experiments}

\input{sections/07_conclusion}
\input{sections/08_limitations}

\bibliography{references}
\appendix
\input{sections/09_appendix}
\end{document}

%% file: sections/01_introduction.tex
\begin{figure}[t]
  \centering
  \includegraphics[width=\columnwidth]{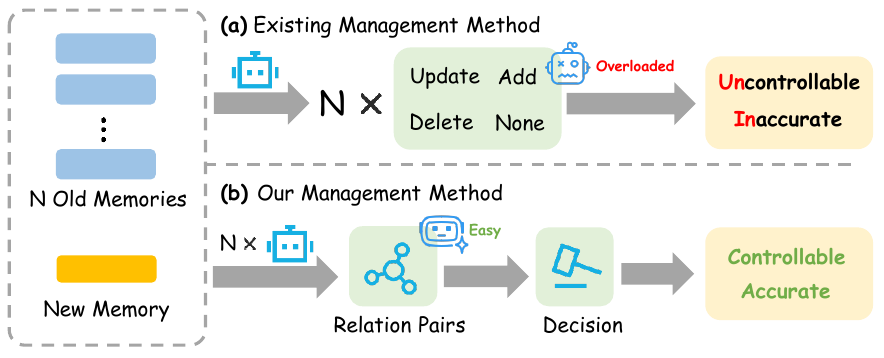} 
  \caption{\textbf{Motivation for ROAM.} Existing methods ask the model to directly choose a memory operation for each old--new memory pair, making the task complex and the results unreliable. ROAM first identifies the semantic relations between memories, then uses them to guide memory organization and fusion, making memory management more controllable and accurate.}
  \label{fig:intro}
\vspace{-10pt}
\end{figure}

\section{Introduction}

Long-lived language-model agents must continuously accumulate user preferences, circumstances, and experiences across sessions while seamlessly revising that knowledge as real-world states change~\citep{sun-etal-2026-preference,zhong2023memorybank}. Because raw conversational histories rapidly exceed an agent's working context limit, contemporary architectures maintain external memory stores to retrieve relevant historical subsets at query time~\citep{zhong2023memorybank,maharana-etal-2024-evaluating,tan-etal-2025-prospect}. To maximize retrieval precision under strict context budgets, systems increasingly represent durable knowledge as fine-grained, atomic facts or structured relations~\citep{modarressi2023retllm,gutierrez2024hipporag,chhikara2025mem0,shen2026anchormem}. Compared to storing full dialogue transcripts, atomic representations preserve key salient information in significantly fewer tokens, provide precise addressable units, and enable targeted updating. Consequently, atomization has become the standard representational foundation for long-term memory systems.

However, atomization specifies only the granularity of memory—it does not govern how accumulating atomic facts should coexist over time as information evolves. Continuous streaming of atomic observations inevitably introduces semantic restatements, spatial or attribute refinements, and temporal state transitions. Merely appending every observation creates severe redundancy and ambiguity, while indiscriminate deletion or overwriting destroys the historical context needed to trace knowledge evolution. Crucially, uncurated, overlapping, and outdated records directly compete with answer-critical evidence for limited retrieval capacity, degrading downstream response quality~\citep{liu-etal-2024-lost,amiraz-etal-2025-distracting}. Existing approaches attempt to maintain growing memory stores through imperative, direct-operation mechanisms (e.g., predicting actions like $\textsc{add}$, $\textsc{update}$, or $\textsc{delete}$)~\citep{chhikara2025mem0}. Yet, such paradigms force the model to infer underlying semantic relations, map them to physical storage mutations, and coordinate updates in a single, black-box compound decision. Conflating semantic interpretation with physical storage execution makes operational consequences unstable and hinders historical traceability. Instead of relying on end-to-end storage mutations, effective atomic memory management requires decoupling explicit semantic relation inference from deterministic state control.

We introduce ROAM (\underline{\textbf{R}}obust \underline{\textbf{O}}rganization of \underline{\textbf{A}}tomic \underline{\textbf{M}}emories for Long-Lived Agents through Semantic Relations), a framework that factorizes this compound decision into explicit relation inference followed by deterministic policy execution. For each incoming observation, ROAM predicts one of five fine-grained semantic relations ($\textsc{ind}$, $\textsc{eqv}$, $\textsc{osn}$, $\textsc{nso}$, $\textsc{con}$) against existing memories. A fixed policy then assigns memories into structural roles—promoting the most informative fact to Primary status while retaining supporting or superseded facts as Evidence. Finally, a fusion module synthesizes compatible details into a compact read view for each Primary entry, allowing downstream retrieval to search exclusively over Primary views. By replacing black-box mutations with relation-guided control, ROAM eliminates redundancy while preserving historical provenance for optimal query answering.

We evaluate ROAM in three complementary settings: a controlled intervention that varies semantic and temporal competition, the original long-horizon histories in LongMemEval~\citep{wu2024longmemeval}, and post-change questions involving multiple entities in MEME-Post~\citep{jung2026meme}. Across three manager models, ROAM achieves the highest mean accuracy at every nonzero level of added competition and on full LongMemEval, while consistently outperforming model-based management baselines on MEME-Post. Relation and fusion ablations show that equivalence, directional containment, conflict, and fused representations provide complementary gains; retrieval diagnostics further link ROAM's robustness to greater coverage of answer-critical memories. Together, these results show that explicit relation-guided organization helps long-term memory systems sustain answer quality as semantic and temporal competition increases.

%% file: sections/03_method.tex
\section{Method}
\label{sec:method}

\begin{figure*}[t]
\vspace{-10pt}
\centering
\includegraphics[width=\textwidth]{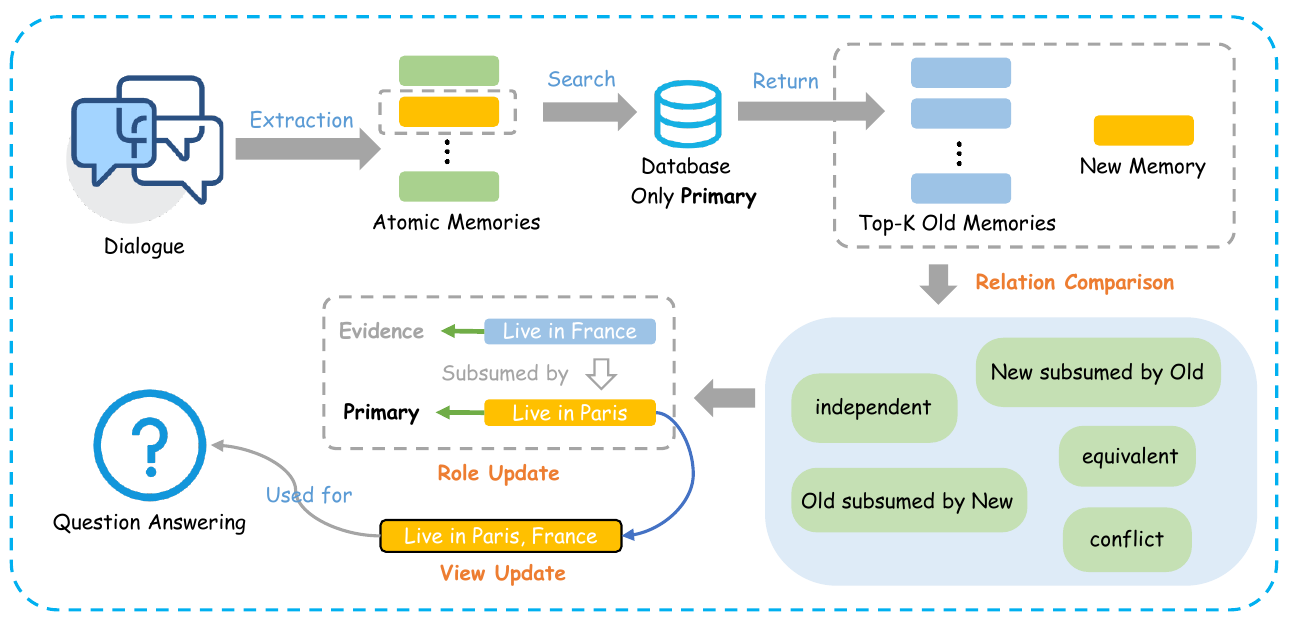}
\caption{\textbf{Overview of the ROAM Framework.} ROAM separates memory management from answer-time retrieval. Incoming atoms are compared with Primary atomic texts and assigned Primary or Evidence roles through five-way relation inference. Fusion constructs a read view for each surviving Primary; answer-time retrieval searches only these views, while Evidence remains preserved but inactive.}
\label{fig:method-overview}
\vspace{-10pt}
\end{figure*}

\subsection{Problem Setup and Scope}

In multi-turn, long-term agent interaction scenarios, the fundamental goal of a memory system is to answer downstream queries by delivering relevant history within a tight context budget $B$. To achieve this, the system operates through a structured three-stage pipeline consisting of \textbf{Extraction}, \textbf{Management}, and \textbf{Retrieval}:
\begin{equation}
\begin{aligned}
\mathcal{F}_t &= \operatorname{Extract}(D_t), \\
\mathcal{S}_t &= \operatorname{Manage}(\mathcal{S}_{t-1}, \mathcal{F}_t), \\
\mathcal{K}_q &= \operatorname{Retrieve}(q, \mathcal{S}_t; B).
\end{aligned}
\end{equation}
Specifically, given raw interaction $D_t$ at turn $t$, the \textbf{Extractor} distills raw text into unstructured factual statements $\mathcal{F}_t$; the \textbf{Manager} updates the structured memory state $\mathcal{S}_{t-1}$ into $\mathcal{S}_t$ by integrating $\mathcal{F}_t$; and the \textbf{Retriever} selects the most relevant subset $\mathcal{K}_q \subseteq \mathcal{S}_t$ to satisfy downstream query $q$ under context budget $B$.

To meet the high-precision retrieval demands of downstream tasks, raw dialogue turns or coarse summaries are insufficient. This scenario naturally mandates representing extracted facts $\mathcal{F}_t$ as fine-grained, \textbf{atomic observations}:
\begin{equation}
O_t = (a_t, \tau_t, c_t),
\end{equation}
where $a_t$ represents a self-contained factual statement, $\tau_t$ is its timestamp, and $c_t$ denotes its source context. Breaking down complex interactions into standalone atomic units provides the necessary fine-grained entry point for flexible memory operations.

However, the streaming influx of atomic observations creates severe challenges for the \textbf{Management} stage. Over time, newly incoming atomic entries frequently repeat known facts, introduce partial details, or invalidate past states. Without rigorous management, these uncurated observations rapidly accumulate, allowing redundant and outdated entries to crowd out valuable knowledge during downstream retrieval. This motivates us to optimize the intermediate \textbf{Management} stage, with the goal of consolidating incoming observations while preserving their historical context and provenance.

\begin{tcolorbox}[breakable,colback=yellow!2,title={\textbf{Remark}}]
In long-term agent interactions, memory management need not occur synchronously after every observation. Atomic observations can be consolidated asynchronously or periodically in the background, allowing the \textbf{Management} stage to prioritize \textbf{memory compactness} over minimal update latency.
\end{tcolorbox}

\subsection{ROAM: Semantic-Relation-Based Memory Management}
Traditional memory management mechanisms commonly operate through imperative storage actions (e.g., $\textsc{add}$, $\textsc{update}$, $\textsc{delete}$). However, applying imperative mutations directly to incoming atomic observations creates a dilemma between accepting redundant storage or suffering permanent information loss from overwriting prior entries~\citep{chhikara2025mem0}.  \textbf{This raises a critical question:} \textit{How can we efficiently consolidate atomic observations to remove redundancy and track knowledge evolution without sacrificing historical provenance or context?} To address this challenge, we propose ROAM, a relation-guided memory management framework that decouples semantic relation modeling from physical storage control. For any incoming observation, ROAM first infers its precise semantic relation against existing memories across five predefined categories. Based on the resolved relation, a deterministic policy reorganizes memories into Primary (active) and Evidence (supporting) entries, preserving historical provenance while preventing redundant accumulation. Finally, ROAM fuses compatible details into a unified read view for each Primary entry, allowing downstream retrieval to query over a compact, highly dense memory state within budget $B$. Figure~\ref{fig:method-overview} provides an overview of the proposed framework, while the detailed streaming procedure is formalized in Algorithm~\ref{alg:roam}.
\subsubsection{Semantic Relations between Atomic Memories}

Let $\mathcal{P}_{t-1}$ be the active atomic memories eligible for management-time retrieval. For an incoming observation $O_t$, ROAM retrieves candidate memories using their atomic text:
\begin{equation}
\mathcal{C}_t=\rho_k(a_t,\mathcal{P}_{t-1}),
\label{eq:management-retrieval}
\end{equation}
where $\rho_k$ returns the $k$ most relevant Primary memories. A relation model then predicts the relation of each ordered pair $(a_t,a_i)$:
\begin{equation}
r_t(i)=\Phi(a_t,a_i)\in\mathcal{R}
\end{equation}
where $\mathcal{R}=\{\textsc{ind},\textsc{eqv},\textsc{osn},
\textsc{nso},\textsc{con}\}$,
\textbf{\textsc{ind} (independent)} means that both observations can be true and neither entails the other;
\textbf{\textsc{eqv} (equivalent)} means that they have the same truth conditions and mutually entail one another;
\textbf{\textsc{osn} (old subsumed by new)} means that the incoming observation strictly entails the old one and adds specificity; for example, ``The user lives in Cambridge, Massachusetts'' entails ``The user lives in Massachusetts''; \textbf{\textsc{nso} (new subsumed by old)} is the reverse entailment direction and adds no specificity;
Finally, \textbf{\textsc{con} (conflict)} means that, after entity, attribute, and temporal scope are aligned, the observations describe incompatible states. Explicit temporal qualifiers take precedence when resolving such conflicts.

\subsubsection{Relation-Conditioned Memory Consolidation}

The predicted relation determines a management action but never directly overwrites a source observation. ROAM stores each observation as an immutable record with a management role and an answer-time read view:
\begin{equation}
M_i=(a_i,v_i,\tau_i,c_i,z_i),
\end{equation}
where $a_i$ is atomic text, $v_i$ is its current read view, and $z_i$ is its role. Atomic text supports relation inference and provenance, whereas the read view provides a compact representation for answering. The two roles are
\begin{equation}
\begin{aligned}
\mathcal{P}_t&=\{M_i:z_i=\textsc{primary}\},\\
\mathcal{E}_t&=\{M_i:z_i=\textsc{evidence}\}.
\end{aligned}
\end{equation}
Primary records remain eligible for subsequent management-time retrieval. Evidence records are retained for provenance but do not compete independently at answer time.

Because different candidates may receive different labels, ROAM selects the highest-priority relation:
\begin{equation}
\textsc{con}\succ\textsc{osn}\succ\textsc{eqv}\succ\textsc{nso}\succ\textsc{ind}.
\label{eq:priority}
\end{equation}
This ordering protects state distinctions first, favors newly introduced specificity next, and consolidates only redundant observations. If $r_t^*$ is the selected relation, its matched candidates are
\begin{equation}
\mathcal{H}_t=\{M_i\in\mathcal{C}_t:r_t(i)=r_t^*\}.
\end{equation}

\begin{table}[t]
\centering
\caption{Relation-conditioned memory updates. Atomic text remains immutable. Role Update denotes the deterministic mapping.}
\label{tab:relation-actions}
\small
\begin{tabular}{p{0.12\columnwidth}p{0.35\columnwidth}p{0.26\columnwidth}}
\toprule
\textbf{Relation} & \textbf{Role update} & \textbf{View update} \\
\midrule
\textsc{ind} & Insert $M_t$ as a new Primary. & $v_t\gets a_t$. \\
\textsc{eqv} & Earliest matched Primary survives; others become Evidence. & Remove redundancy. \\
\textsc{osn} & $M_t$ becomes Primary; old matches become Evidence. & Preserve added specificity. \\
\textsc{nso} & Earliest matched Primary survives; $M_t$ becomes Evidence. & Preserve existing information. \\
\textsc{con} & $M_t$ becomes current Primary; old matches become Evidence. & Build temporal history. \\
\bottomrule
\end{tabular}
\end{table}

Thus, \textsc{eqv} and \textsc{nso} preserve the earliest anchor, whereas \textsc{osn} and \textsc{con} promote the incoming observation; \textsc{ind} creates a new Primary. For every non-independent update, let $M^*$ be the surviving Primary and let $\mathcal{D}_t$ be the records assigned to Evidence. Sorting $\{M^*\}\cup\mathcal{D}_t$ by observation time as $(S_1,\ldots,S_n)$, ROAM updates its read view by
\begin{equation}
\begin{aligned}
v_{M^*}&\gets\operatorname{Fuse}\bigl(
\operatorname{View}(S_1),\ldots,\\
&\qquad\operatorname{View}(S_n);r_t^*\bigr).
\end{aligned}
\label{eq:fusion}
\end{equation}
Fusion changes only the read view: it removes redundancy for \textsc{eqv}, preserves compatible specificity for \textsc{osn}/\textsc{nso}, and records temporal state changes for \textsc{con}. Answer-time retrieval uses Primary views, while subsequent management continues to use immutable atomic texts.
\begin{algorithm}[t]
\caption{ROAM Atomic Memory Update}
\label{alg:roam}
\begin{algorithmic}[1]
\REQUIRE New record $M_t=(a_t,a_t,\tau_t,c_t,\cdot)$ and state $(\mathcal{P},\mathcal{E})$
\ENSURE Updated state and Primary views
\STATE $\mathcal{C}\gets\rho(a_t,\{a_i:M_i\in\mathcal{P}\},k)$
\IF{$\mathcal{C}=\emptyset$}
    \STATE $r^*\gets\textsc{ind}$
\ELSE
    \STATE $\mathcal{L}\gets\{\Phi(a_t,a_i):M_i\in\mathcal{C}\}$
    \STATE $r^*\gets\operatorname{HighestPriority}(\mathcal{L})$
\ENDIF
\IF{$r^*=\textsc{ind}$}
    \STATE Insert $M_t$ as a new Primary
\ELSE
    \STATE $\mathcal{H}\gets\operatorname{Matches}(\mathcal{C},r^*)$
    \STATE $(M^*,\mathcal{D})\gets\operatorname{RoleUpdate}_{r^*}(\mathcal{H},M_t)$
    \STATE Update roles and sets: $M^*$ Primary, $\mathcal{D}$ Evidence
    \STATE Sort $\{M^*\}\cup\mathcal{D}$ by time and form $\mathbf{V}$
    \STATE $v_{M^*}\gets\operatorname{FuseOrdered}(\mathbf{V};r^*)$
\ENDIF
\end{algorithmic}
\end{algorithm}
\vspace{-5pt}

%% file: sections/04_experiments.tex
\section{Experiments}
\label{sec:experiments}

Section~\ref{sec:method} introduced ROAM, a relation conditioned memory management method. By analyzing the relations between incoming and stored memories, ROAM organizes, consolidates, and updates candidate memories to construct a compact, query-oriented collection of retrievable memory units. We refer to this collection as the \emph{active memory set}: the memory units that are eligible for retrieval at answer time.

This section evaluates whether ROAM remains robust under retrieval-budget interference, whether its gains transfer to natural long-horizon and state-update tasks, and which relations and system components account for its gains.

\subsection{Experimental Setup}

\subsubsection{Benchmarks}

We use three complementary evaluation settings:

\textbf{LongMemEval}~\citep{wu2024longmemeval}.
We evaluate the unmodified 500-question LongMemEval benchmark to test transfer to natural long-horizon conversational histories. Memory histories, question difficulty, and candidate-memory distributions are all determined by the original dataset.

\textbf{MEME-Post}.
MEME~\citep{jung2026meme} involves multiple entities and values that may remain stable or change over time. We process preceding sessions chronologically and evaluate all 694 post-change questions using the final-state question-answering formulation adopted throughout this study. We call the resulting unfiltered question-level accuracy \emph{MEME-Post}; this setting tests transfer to multi-entity memory and state-update scenarios.

\textbf{LongMemEval Controlled.} 
LongMemEval and MEME test whether a complete memory system can answer questions from long conversational histories, but their fixed histories, difficulty, and candidate-memory distributions give end-to-end accuracy limited leverage for analyzing how systems organize redundant or unhelpful content and correctly resolve changes between old and new states.
To address this, we construct LongMemEval Controlled from the 470 LongMemEval questions with usable gold-memory annotations. The core design holds the answer-critical memories fixed for each question and varies only the number of additional confounders a management policy must handle, evaluating at $N \in \{0, 2, 4, 6, 8\}$. This setting provides a precise diagnosis of whether a memory-management strategy can effectively suppress semantic interference and accurately retrieve critical memories under a constrained retrieval budget. The full construction protocol is given in Appendix~\ref{app:lme-controlled}.

\begin{table*}[t]
\centering
\caption{Answer accuracy (\%; higher is better) with a 256-token memory budget. Because Append-all is manager-independent, its result is repeated across manager-model blocks. \textbf{Boldface} and \underline{underlining} mark the highest and second-highest means, respectively, within each manager-model block and evaluation condition.}
\label{tab:main-results}
\resizebox{\textwidth}{!}{%
\begin{tabular}{llrrrrrrr}
\toprule
\multirow{2}{*}{\textbf{Manager model}} & \multirow{2}{*}{\textbf{Method}} & \multicolumn{5}{c}{\shortstack{\textbf{LongMemEval Controlled}}} & \multirow{2}{*}{\shortstack{\textbf{LongMemEval}}} & \multirow{2}{*}{\shortstack{\textbf{MEME-Post}}} \\
\cmidrule(lr){3-7}
 & & \textbf{$N=0$} & \textbf{$N=2$} & \textbf{$N=4$} & \textbf{$N=6$} & \textbf{$N=8$} & & \\
\midrule
\multirow{4}{*}{\shortstack{\texttt{Gemma-4-}\texttt{12B}}} & Append-all & \underline{87.0} & \underline{77.3} & 47.0 & 25.4 & 24.4 & \underline{57.1} & \underline{32.5} \\
 & Mem0 & 74.3 & 72.3 & \underline{74.0} & \underline{71.7} & \underline{69.1} & 55.0 & 24.8 \\
 & EverMemOS-style & 86.2 & 74.5 & 48.5 & 38.3 & 36.0 & 54.8 & 32.4 \\
 & \textbf{ROAM (Ours)} & \textbf{87.2} & \textbf{81.5} & \textbf{78.3} & \textbf{76.2} & \textbf{71.3} & \textbf{63.0} & \textbf{38.9} \\
\midrule
\multirow{4}{*}{\shortstack{\texttt{Qwen3.5-}\texttt{9B}}} & Append-all & \textbf{87.0} & \underline{77.3} & 47.0 & 25.4 & 24.4 & 57.1 & \underline{32.5} \\
 & Mem0 & 85.3 & 75.7 & \underline{67.0} & \underline{54.7} & \underline{42.1} & \underline{57.8} & 31.4 \\
 & EverMemOS-style & \underline{86.0} & 74.0 & 47.2 & 37.2 & 36.0 & 55.8 & 24.9 \\
 & \textbf{ROAM (Ours)} & 84.9 & \textbf{78.1} & \textbf{78.3} & \textbf{76.6} & \textbf{71.9} & \textbf{59.8} & \textbf{33.1} \\
\midrule
\multirow{4}{*}{\shortstack{\texttt{Qwen3-}\texttt{8B}}} & Append-all & \textbf{87.0} & \underline{77.3} & 47.0 & 25.4 & 24.4 & \underline{57.1} & \textbf{32.5} \\
 & Mem0 & 76.8 & 64.0 & \underline{59.8} & \underline{53.2} & \underline{48.9} & 54.2 & 30.3 \\
 & EverMemOS-style & \underline{86.2} & 73.2 & 48.3 & 39.6 & 37.0 & 54.4 & 23.6 \\
 & \textbf{ROAM (Ours)} & 84.7 & \textbf{77.9} & \textbf{74.9} & \textbf{68.1} & \textbf{64.3} & \textbf{58.6} & \underline{31.6} \\
\bottomrule
\end{tabular}
}%
\end{table*}

\subsubsection{Baselines}

We compare three memory-management policies spanning no active management, model-driven updates, and semantic consolidation:
\begin{itemize}[leftmargin=1em,itemsep=0pt,label=$\triangleright$,topsep=0pt]
    \item \textbf{Append-all} retains every candidate as an independently retrievable entry without active consolidation or deletion. Therefore, this method does not require a manager model.
    \item \textbf{Mem0}~\citep{chhikara2025mem0} uses the manager to choose \textsc{add}, \textsc{update}, \textsc{delete}, or \textsc{none} for each incoming memory and retrieved history.
    \item \textbf{EverMemOS-style}~\citep{hu2026evermemos} applies semantic grouping and consolidation only to short factual memories, excluding the full system's trace formation, profile building, and agentic recollection.
\end{itemize}
\subsection{Implementation Details}
The main experiments use \texttt{Qwen3.5-\allowbreak 9B}~\citep{qwen3.5}, \texttt{Qwen3-\allowbreak 8B}~\citep{yang2025qwen3}, and \texttt{Gemma-4-\allowbreak 12B}~\citep{gemmateam2026gemma4} as manager models for cross-family comparisons. To analyze the effect of parameter scale, we additionally compare \texttt{Gemma-4-E4B}, \texttt{Gemma-4-12B}, and \texttt{Gemma-4-26B-A4B} within the same model family. Within each comparison, all managed methods use the same manager model. The relation, fusion, budget, and main-text retrieval analyses use \texttt{Qwen3.5-9B}; the supplementary retrieval breakdown additionally reports results using \texttt{Gemma-4-12B}.

Across all configurations, methods receive the same extracted candidate-memory stream and initial store. We use \texttt{Qwen3-\allowbreak Embedding-\allowbreak 0.6B}~\citep{zhang2025qwen3embedding} for retrieval, \texttt{Gemma-4-\allowbreak 26B-\allowbreak A4B} for answer generation, and \texttt{DeepSeek-V4-\allowbreak Flash}~\citep{deepseekai2026deepseekv4} for evaluation. All methods operate under the same 256-token memory-context budget. For ROAM, the shared retriever embeds Primary atomic texts during memory management and Primary read views during answering. Consequently, performance differences primarily reflect how each method organizes memories and constructs retrieval units for answering.

Table~\ref{tab:main-results} summarizes answer accuracy on LongMemEval Controlled, full LongMemEval, and MEME-Post for the three manager models.

\subsection{Main Results}

\paragraph{Robustness to memory interference and consistency across manager models.}
ROAM ranks first across all non-zero interference settings, and its advantage remains consistent across all three manager models. At $N=8$, it surpasses the strongest baseline in each group by up to 29.8 percentage points, with substantially smaller degradation as interference increases. This indicates that the gains stem from the relation-conditioned organization policy itself rather than reliance on a specific manager model. In contrast, Append-all reaches 87.0\% under $N=0$ but plummets to 24.4\% at $N=8$, showing that retaining all candidates is effective only when the memory store is uncontested, whereas explicitly controlling the active memory set yields stable advantages under interference.

\paragraph{Generalization across evaluation settings.}
The advantage also transfers to natural scenarios. On the full LongMemEval, ROAM achieves the highest accuracy under all three manager models (63.0\%, 59.8\%, and 58.6\%), exceeding the corresponding strongest baselines by 5.9, 2.0, and 1.5 percentage points. On MEME-Post, ROAM outperforms model-based baselines with every manager and ranks first overall with \texttt{Gemma-4-12B} and \texttt{Qwen3.5-9B} (38.9\% and 33.1\%). With \texttt{Qwen3-8B}, ROAM remains the strongest managed store at 31.6\%. These results demonstrate that ROAM scales to long-horizon histories, multi-entity interactions, and post-update questions.

\begin{table}[t]
\centering
\caption{Relation and fusion ablations with \texttt{Qwen3.5-9B} and a 256-token memory budget. Values are answer accuracy (\%; higher is better). Append-all and ROAM reuse the corresponding results from Table~\ref{tab:main-results}; other rows use matched ablation runs. \textbf{Boldface} and \underline{underlining} mark the highest and second-highest values, respectively, at each interference level.}
\label{tab:relation-ablation}
\footnotesize
\setlength{\tabcolsep}{3pt}
\resizebox{\columnwidth}{!}{%
\begin{tabular}{lccccc}
\toprule
\textbf{Management policy} & \textbf{$N=0$} & \textbf{$N=2$} & \textbf{$N=4$} & \textbf{$N=6$} & \textbf{$N=8$} \\
\midrule
Append-all & \textbf{87.0} & 77.3 & 47.0 & 25.4 & 24.4 \\
CON-only & 84.0 & 77.2 & 54.3 & 34.9 & 32.1 \\
CON+EQV & 84.0 & \textbf{78.3} & 72.8 & 66.6 & 62.3 \\
ROAM w/o Fusion & 79.1 & 77.9 & \underline{75.5} & \underline{73.6} & \underline{70.9} \\
ROAM & \underline{84.9} & \underline{78.1} & \textbf{78.3} & \textbf{76.6} & \textbf{71.9} \\
\bottomrule
\end{tabular}
}%
\end{table}
\subsection{Ablation Studies}

The main results establish that ROAM is most effective when the memory store contains many plausible candidates. We therefore examine which design choices in its relation-aware memory policy account for this robustness.

\subsubsection{Which Relations Matter?}

We compare three relation policies with progressively greater expressive power. \textbf{CON-only} handles contradiction relations. \textbf{CON+EQV} additionally handles equivalence relations. Full \textbf{ROAM} further models directional containment through the \textsc{osn} and \textsc{nso} relation types. Table~\ref{tab:relation-ablation} evaluates each policy across the interference sweep.

The overall trend is clear: as the policy covers more relation types, performance becomes more stable as interference increases. At $N=8$, CON-only, CON+EQV, and ROAM achieve 32.1\%, 62.3\%, and 71.9\%, respectively. From $N=4$ onward, their ordering is consistent:
$
\textbf{\text{CON-only} < \text{CON+EQV} < \text{ROAM}.}
$
These results show that contradiction, equivalence, and directional-containment relations all contribute to memory management. Each captures a different form of interaction among memories, and together they enable ROAM to construct a more reliable active set when candidate memories compete for limited context.

\subsubsection{Effect of Fusion}

Relation ablations confirm that richer relation modeling improves the active set. To test whether the answer-time representation matters independently, we construct \textbf{ROAM w/o Fusion}: it keeps the same managed candidates, relation predictions, and Primary--Evidence assignments as ROAM, but retrieves the original Primary text instead of generating fused views. This isolates representation effects while holding active-set management fixed.

Table~\ref{tab:relation-ablation} suggests that Fusion improves accuracy at every interference level, so its benefit is consistent across all interference strengths rather than limited to a specific regime. Relation-conditioned management alone already provides a substantial gain, and fused views further increase accuracy. Because the two variants share identical relation predictions and Primary--Evidence assignments, the comparison shows that the mechanisms are complementary: management selects what to retain, while fusion determines how that information is organized and presented to the answer model.

\subsection{Analysis}

In this section, we want to address two questions. First, does ROAM remain effective across manager-model scales and memory-context budgets? Second, do its relation decisions improve the final context through the intended mechanism?  

\subsubsection{Manager-Model Scale Sensitivity}

\begin{figure}[t]
\centering
\includegraphics[width=\columnwidth]{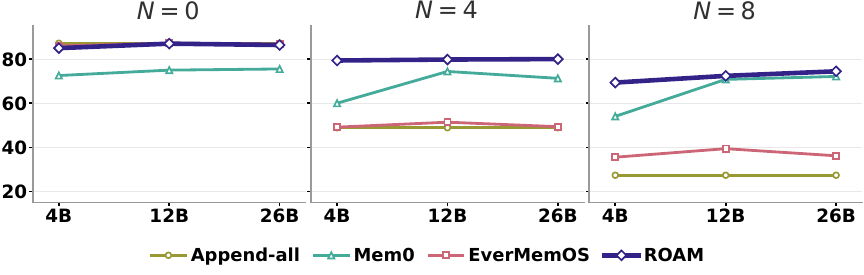}
\caption{Manager-model scale sensitivity. The horizontal axis varies the \texttt{Gemma-4} manager from \texttt{E4B} to \texttt{26B-A4B}; the vertical axis reports answer accuracy. Panels show $N\in\{0,4,8\}$, with all other components fixed.}
\label{fig:manager-size-sensitivity}
\end{figure}

Because the manager model performs relation inference, its capacity may affect the quality of retrieval-eligibility decisions. Figure~\ref{fig:manager-size-sensitivity} isolates this factor within the \texttt{Gemma-4} family while holding all other components fixed. ROAM remains consistently strong across \texttt{E4B}, \texttt{12B}, and \texttt{26B-A4B}, ranking first at all displayed nonzero interference levels for all three scales; its advantage therefore does not depend on a particular manager-model size.

We further find that additional capacity becomes most useful as competition intensifies. At N=8, ROAM improves from 69.4\% with \texttt{E4B} to 71.3\% with \texttt{12B} and 74.5\% with \texttt{26B-A4B}, a 5.1-point gain overall. By contrast, the three variants perform more similarly under low interference, where few candidates compete for the available context. This interaction is consistent with ROAM's design: stronger relation inference matters most when the active memory set must preserve answer-relevant information under heavy competition.

\subsubsection{Memory-Context Budget Sensitivity}

\begin{figure}[t]
\centering
\includegraphics[width=\columnwidth]{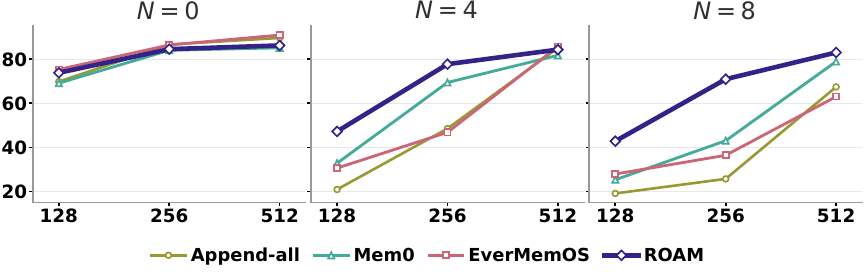}
\caption{Memory-context budget sensitivity with \texttt{Qwen3.5-9B}. The horizontal axis gives the memory-context budget in tokens; the vertical axis reports answer accuracy . Panels show $N\in\{0,4,8\}$.}
\label{fig:token-budget-sensitivity}
\vspace{-10pt}
\end{figure}

The memory-context budget determines how selectively those decisions must allocate retrieval capacity. Figure~\ref{fig:token-budget-sensitivity} examines this complementary factor by fixing \texttt{Qwen3.5-9B} as the manager and varying the budget over 128, 256, and 512 tokens.

ROAM's advantage is most pronounced when the context is constrained or interference is strong. At $N=8$, it leads at all three budgets, reaching 42.8\%, 71.9\%, and 83.0\% and outperforming the strongest baseline by 14.9, 29.8, and 4.1 points, respectively. Thus, even as the budget increases fourfold, ROAM retains its advantage in the most competitive setting. These results indicate that the gain comes from organizing limited context among plausible candidates rather than merely increasing the amount of available context.

\subsubsection{Relation-Inference Quality}

We next examine the internal control signal that supports the robustness of ROAM. ROAM formulates memory management as five-way relation inference and uses the predicted relation to assign Primary--Evidence roles and retrieval eligibility. Reliable relation inference is therefore a prerequisite for constructing the intended active memory set. We evaluate \texttt{Gemma-4-26B-A4B-it} on 843 memory pairs sampled from PersonaMem-v2 dialogues~\citep{jiang2025personamemv2}. \texttt{DeepSeek-V4-Pro} provides initial labels for stratified sampling across the five relations, after which we manually verify every pair against ROAM's definitions. This construction provides sufficient support for a class-balanced assessment of relation discriminability.

As Table~\ref{tab:relation-inference-main} shows, \texttt{Gemma-4-26B-A4B-it} identifies equivalence and conflict particularly well, with F1 scores of 93.16\% and 91.02\%. It also distinguishes the two directional subsumption relations, achieving 76.08\% for \textsc{osn}, where the incoming memory is more specific, and 69.37\% for \textsc{nso}, where the stored memory is more specific. Overall, the results show that ROAM's five-way relation space provides an explicit and measurable interface for memory management.

\begin{table}[t]
\centering
\caption{Relation-inference performance (\%) of \texttt{Gemma-4-26B-A4B-it} on 843 manually verified PersonaMem-v2 memory pairs. Stratified sampling supports balanced evaluation across the five relations.}
\label{tab:relation-inference-main}
\small
\setlength{\tabcolsep}{4pt}
\resizebox{0.4\textwidth}{!}{
\begin{tabular}{lrrr}
\toprule
\textbf{Class} & \textbf{Precision} & \textbf{Recall} & \textbf{F1} \\
\midrule
\textsc{ind} & 65.5 & 96.5 & 78.0 \\
\textsc{eqv} & 97.2 & 89.3 & \textbf{93.1} \\
\textsc{osn} & 85.4 & 68.5 & 76.0 \\
\textsc{nso} & 65.7 & 73.4 & 69.3 \\
\textsc{con} & 98.0 & 84.9 & 91.0 \\
\midrule

Macro avg. & 82.4 & 82.5 & 81.5 \\
\bottomrule
\end{tabular}}
\vspace{-10pt}
\end{table}

\subsubsection{Retrieval Mechanism Analysis}

Reliable relation predictions matter only if the resulting management decisions improve the context presented to the answer model. We therefore inspect the final 256-token context at $N=8$ with \texttt{Qwen3.5-9B}. We report four complementary metrics. \emph{Any-source coverage} asks whether the context represents at least one required gold source, while \emph{all-source coverage} requires every gold source. \emph{Gold-source recall} measures the fraction of required sources represented per query, and \emph{explicit confounder-token share} measures the fraction of retained tokens uniquely attributable to confounders. Coverage propagates deduplicated provenance from atomic records to Primary read views that contribute tokens to the final context.

Table~\ref{tab:retrieval-metrics} shows that ROAM provides the strongest coverage on all three gold-source metrics. 
It represents at least one required source for 95.1\% of queries and all required sources for 73.8\%, with 84.6\% gold-source recall. 

All methods use almost the full budget, so the coverage gain does not come from exposing the answer model to more context. ROAM instead achieves higher coverage with 4.7 retrieved Primary read views on average, compared with 4.8--5.1 units for the baselines. The mechanism is therefore selective context organization: ROAM allocates more of the same context budget to answer-critical sources while reducing competition from confounders.

\begin{table}[t]
\centering
\caption{retrieval diagnostics with \texttt{Qwen3.5-9B} at $N=8$ under a 256-token memory-context budget. Gold-source metrics use deduplicated provenance; an original entry or its provenance-linked Primary read view covers the corresponding source. Explicit confounder tokens exclude mixed or unattributed tokens. Ever-s. denotes the adapted EverMemOS-style baseline.}
\label{tab:retrieval-metrics}
\scriptsize
\setlength{\tabcolsep}{3pt}
\resizebox{0.48\textwidth}{!}{
\begin{tabular}{lrrrr}
\toprule
\textbf{Metric} & \textbf{Append} & \textbf{Mem0} & \textbf{Ever-s.} & \textbf{ROAM} \\
\midrule
Any-source cov. (\%) $\uparrow$ & 36.6 & 81.3 & 35.5 & \textbf{95.1} \\
All-source cov. (\%) $\uparrow$ & 21.7 & 57.4 & 18.5 & \textbf{73.8} \\
Gold-source recall (\%) $\uparrow$ & 29.1 & 69.0 & 26.9 & \textbf{84.6} \\
Confounder tokens (\%) $\downarrow$ & 84.0 & 58.6 & 68.3 & \textbf{47.1} \\
\bottomrule
\end{tabular}}
\end{table}

%% file: sections/07_conclusion.tex
\section{Conclusion}

We introduced ROAM, a relation-guided framework for managing atomic factual memories under finite retrieval budgets. ROAM separates constrained semantic relation inference from deterministic state control, retaining immutable observations in auditable Primary--Evidence structures while exposing compact fused views for retrieval.

Across controlled interference, full LongMemEval, and MEME-Post, ROAM consistently outperforms model-based memory-management baselines across three manager models. Ablations and retrieval diagnostics show that relation-guided control of the active memory set drives most of the gain by improving coverage of answer-critical evidence and reducing competition from redundant or superseded records, while fusion provides a complementary benefit. These results suggest that long-term agent memory should decouple source retention from retrieval eligibility, preserving historical evidence without requiring every observation to compete for limited answer-time context.

%% file: sections/08_limitations.tex
\section*{Limitations}

Our evaluation uses benchmark histories and controlled retrieval budgets, which enable measurable interference and consistent comparisons but cannot capture all deployments. Applications may differ in stream length, update frequency, context limits, and conversational complexity; gains should therefore be interpreted within the evaluated settings.

ROAM also operates on short, extracted factual units. This makes relation inference and provenance explicit, but leaves hierarchical, event-level, and richer structured memories unexplored. Extending the Primary--Evidence organization and active-set retrieval principles to these representations is an important direction for future work.

%% file: sections/09_appendix.tex
\input{sections/02_related_work}

\section{LLM Usage Statement}
\section*{LLM Usage}

LLMs were used both as components of the evaluated systems and as tools in the research workflow. In the research workflow, LLMs were used to generate the fixed pool of controlled confounders, provide initial labels for constructing the relation diagnostic, and serve as answer judges. All LLM-generated labels were manually reviewed and corrected. The manager and answer models used in each experiment are specified in the experimental setup.

LLMs were also used solely to improve the language and readability of the manuscript. They did not contribute to the research questions, methodological design, experimental decisions, or interpretation of results. All LLM-assisted edits and research artifacts were reviewed and approved by the authors, who take full responsibility for the final manuscript, its claims, and the released artifacts.

\section{Ethics Statement}
This work studies personal factual memory in language-model agents. The experiments use existing benchmarks and involve neither new human interactions nor the collection of new personal data. Deployed memory systems may nevertheless
expose sensitive information, retain outdated or incorrect claims, and influence later responses. ROAM's Evidence records preserve source history and support auditability but
are not a privacy or security mechanism. Deployments should therefore include informed consent, access control, secure storage, appropriate retention and deletion policies, and mechanisms for users to inspect and correct stored
information. ROAM does not address these broader deployment requirements.

\section{Evaluation Validation}

\paragraph{Answer-judge agreement.}
\label{sec:appendix-judge-validation}

We validate the automated answer judge through a stratified audit of 200
examples spanning all three benchmarks and all four compared memory-management
methods. A human annotator independently assesses the semantic correctness of
each candidate answer with respect to its question and reference answer. The
resulting human judgments agree with the DeepSeek~V4~Flash judge on 97.5\% of
the audited examples. This validation uses the same judge model and prompt as
the main experiments.

\paragraph{Relation-label verification.}
\label{sec:appendix-relation-verification}

The relation diagnostic reported in the main paper contains 843 memory pairs
selected through stratified sampling to obtain approximately balanced coverage
of \textsc{ind}, \textsc{eqv}, \textsc{osn}, \textsc{nso}, and \textsc{con}.
A human annotator reviews every pair against ROAM's relation definitions before
the diagnostic evaluation is performed.

\section{Question-Type Results}
\label{sec:appendix-question-type-results}

Tables~\ref{tab:lme-original-by-type} and~\ref{tab:meme-post-by-type}
report the question-type breakdowns for the full LongMemEval and MEME-Post
evaluations. Append-all does not use a manager model; we therefore average its
two recorded manager-block values for each question type and repeat the pooled
result in both blocks. All values are rounded to one decimal place.

\begin{table*}[t]
\caption{Answer accuracy (\%) by question type on the original LongMemEval benchmark. Temp., Multi., Update, User, Pref., and Asst. denote temporal reasoning, multi-session, knowledge update, single-session user, single-session preference, and single-session assistant questions, respectively. Append-all is pooled across manager-model blocks. Boldface and underlining mark the highest and second-highest values, respectively, within each manager-model block and question type.}
\label{tab:lme-original-by-type}
\centering
\small
\setlength{\tabcolsep}{4pt}
\begin{tabular}{llrrrrrr}
\toprule
\textbf{Manager model} & \textbf{Method} & \textbf{Temp.} & \textbf{Multi.} & \textbf{Update} & \textbf{User} & \textbf{Pref.} & \textbf{Asst.} \\
\midrule
\multirow{4}{*}{\shortstack{Gemma~4\\12B}}
 & Append-all & \underline{56.7} & 36.8 & \underline{83.3} & 93.5 & 83.3 & \textbf{15.2} \\
 & Mem0 & 55.9 & 37.2 & 69.4 & 90.6 & \underline{86.7} & \underline{14.3} \\
 & EverMemOS-style & 49.6 & \underline{39.7} & 73.6 & \underline{93.7} & 86.2 & 12.5 \\
 & \textbf{ROAM (ours)} & \textbf{63.8} & \textbf{43.8} & \textbf{87.5} & \textbf{95.3} & \textbf{90.0} & \underline{14.3} \\
\midrule
\multirow{4}{*}{\shortstack{Qwen3.5\\9B}}
 & Append-all & 56.7 & 36.8 & \textbf{83.3} & \underline{93.5} & 83.3 & \underline{15.2} \\
 & Mem0 & \underline{57.5} & \underline{42.2} & 75.0 & \textbf{95.3} & \underline{86.1} & \textbf{16.1} \\
 & EverMemOS-style & 51.2 & 37.2 & 76.4 & 92.2 & 83.3 & 10.7 \\
 & \textbf{ROAM (ours)} & \textbf{59.1} & \textbf{42.5} & \underline{81.9} & 90.6 & \textbf{86.7} & 14.3 \\
\bottomrule
\end{tabular}
\end{table*}

\begin{table*}[t]
\caption{Answer accuracy (\%) by question type on MEME-Post. The benchmark's question-type abbreviations are retained from the evaluation data. Append-all is pooled across manager-model blocks. Boldface and underlining mark the highest and second-highest values, respectively, within each manager-model block and question type.}
\label{tab:meme-post-by-type}
\centering
\small
\setlength{\tabcolsep}{5pt}
\begin{tabular}{llrrrrrr}
\toprule
\textbf{Manager model} & \textbf{Method} & \textbf{Cas} & \textbf{Abs} & \textbf{Tr} & \textbf{Del} & \textbf{Agg} & \textbf{ER} \\
\midrule
\multirow{4}{*}{\shortstack{Gemma~4\\12B}}
 & Append-all & 18.9 & \underline{9.2} & \underline{64.5} & \underline{12.5} & 15.4 & \textbf{98.1} \\
 & Mem0 & \underline{26.2} & 8.5 & 4.5 & 3.7 & 14.2 & 97.3 \\
 & EverMemOS-style & 15.9 & 7.7 & 61.4 & 11.3 & \textbf{20.0} & \underline{97.6} \\
 & \textbf{ROAM (ours)} & \textbf{28.1} & \textbf{10.3} & \textbf{79.5} & \textbf{20.1} & \underline{15.8} & \underline{97.6} \\
\midrule
\multirow{4}{*}{\shortstack{Qwen3.5\\9B}}
 & Append-all & 18.9 & \textbf{9.2} & \textbf{64.5} & \underline{12.5} & \underline{15.4} & \textbf{98.1} \\
 & Mem0 & 13.4 & \textbf{9.2} & 16.6 & 7.4 & \textbf{19.3} & 97.3 \\
 & EverMemOS-style & \underline{19.5} & 7.7 & 58.3 & 10.8 & 13.2 & 97.3 \\
 & \textbf{ROAM (ours)} & \textbf{23.8} & \underline{8.5} & \underline{60.9} & \textbf{16.6} & 14.7 & \underline{97.6} \\
\bottomrule
\end{tabular}
\end{table*}

\section{LongMemEval Controlled}
\label{app:lme-controlled}

\subsection{Construction and Confounder Audit}
\label{sec:appendix-confounders}

For each of the 470 LongMemEval questions with usable gold-memory annotations, we retain the answer-critical memories and the original
non-gold background memories, then add a controlled number of additional
memories, which we refer to as \emph{confounders}. The answer-critical memories
remain fixed across conditions; only the number and type of confounders vary.
Let $N$ denote the number of confounders in a question's store. We evaluate
$N\in\{0,2,4,6,8\}$.

We construct two types of confounders:
\begin{itemize}[leftmargin=1em,itemsep=0pt,topsep=0pt]
    \item \textbf{Type~I} memories are semantically similar to the query but do
    not answer it: they provide neither the correct answer nor a verifiably
    incorrect alternative.
    \item \textbf{Type~II} memories are used only for Knowledge Update
    questions. They state a valid but temporally obsolete value and conflict
    with the current one.
\end{itemize}

During memory ingestion, methods receive neither the current query,
gold-memory annotations, confounder-type labels, nor the expected answer.
\texttt{Gemma-4-26B-A4B}
generates one fixed pool of eight confounders for each question. The same pool
is then used by every method and manager, independently of their predictions
or outputs.

All generated confounders pass an automated filtering stage before inclusion.
For Type~II confounders, the temporally obsolete value is always assigned an
earlier timestamp than the corresponding current gold value. This preserves
the intended chronological ordering and ensures that Type~II examples
represent genuine temporal supersession.

The resulting controlled dataset contains 3,760 fixed confounders: 3,184
Type~I memories for 398 non-update questions and 576 Type~II memories for 72
Knowledge Update questions. Using Qwen3-Embedding-0.6B and the gold
annotations, we compute the similarity of every confounder to the query and to
each answer-critical memory.
As Table~\ref{tab:confounder-audit} shows, most confounders are more similar to
the query than all required gold memories. The controlled setting therefore
requires systems to handle plausible candidate memories rather than merely
filter obviously irrelevant ones.

\begin{table}[t]
\caption{Audit of the fixed confounder set for the 470 controlled questions.
Type~I memories are semantic and non-answering; Type~II memories contain
temporally obsolete values. Similarity rows report candidate-level means. The
final two rows give the percentage of confounders whose query similarity
exceeds that of at least one or all required gold memories.}
\label{tab:confounder-audit}
\centering
\small
\begin{tabular}{lrr}
\toprule
\textbf{Statistic} & \textbf{Type I} & \textbf{Type II} \\
\midrule
Queries & 398 & 72 \\
Confounders & 3,184 & 576 \\
Confounder--query similarity & 0.7149 & 0.7641 \\
Gold--query similarity & 0.6321 & 0.6926 \\
Above at least one gold (\%) & 76.35 & 83.33 \\
Above all gold (\%) & 62.19 & 73.09 \\
\bottomrule
\end{tabular}
\end{table}

\subsection{Retrieval Diagnostics}
\label{sec:appendix-retrieval}

We compute retrieval diagnostics for Gemma~4~12B and Qwen3.5-9B at $N=8$ over the same 470 questions and three runs used for controlled answer accuracy. Once a run's final store, embeddings, and retrieval order are fixed, the diagnostic is deterministic and requires neither the answer model nor the judge.

\paragraph{Gold-source coverage.}
For each query $q$, let $G_q$ contain stable keys for the required gold sources, and let $S_q$ contain the source keys represented by retrieval units that contribute tokens to the final hard-truncated context. For ROAM, these units are the views attached to retrieved Primaries. We compute
\begin{align}
\mathrm{AnyGold}(q) &= \mathbb{I}[|G_q \cap S_q| > 0], \\
\mathrm{AllGold}(q) &= \mathbb{I}[G_q \subseteq S_q], \\
\mathrm{GoldRecall}(q) &= \frac{|G_q \cap S_q|}{|G_q|}.
\end{align}
Metrics are macro-averaged over queries after deduplicating source keys and retrieved matches. Append-all, EverMemOS-style, and ROAM use exact matches from gold text to ingestion records; ROAM propagates membership into fused views through \texttt{fused\_member\_ids}. Because Mem0 may rewrite text, matching first uses exact identity and then a \texttt{SequenceMatcher} ratio of at least 0.65. Unmapped sources remain in $G_q$ and count as uncovered. These metrics quantify the representation of answer-critical sources in the final context.

\paragraph{Token attribution and truncation.}
The retrieved memory block is serialized and hard-truncated to 256 Qwen3-8B tokenizer tokens. A unit is counted only if it contributes retained tokens; for ROAM, this unit is a Primary view rather than its underlying atoms. Let $T_q$ be the retained memory-context tokens and $C_q$ the subset uniquely attributable to fixed confounder sources. We report the macro-average of $C_q/T_q$; tokens with mixed or unavailable attribution are excluded from the numerator. Formatting, system-prompt, question, and answer-instruction tokens are excluded from $T_q$.

Append-all does not use a manager model, so its retrieval diagnostics are shared across manager-model blocks.

\begin{table}[t]
\caption{Gold-source recall (\%) by confounder type at $N=8$. Type I contains semantic non-answering confounders for 398 queries; Type II contains temporally obsolete values for 72 Knowledge Update queries. Append-all is pooled across manager-model blocks.}
\label{tab:retrieval-by-confounder-type}
\centering
\small
\setlength{\tabcolsep}{4pt}
\begin{tabular}{llrr}
\toprule
\textbf{Manager model} & \textbf{Method} & \textbf{Type I} & \textbf{Type II} \\
\midrule
\multirow{4}{*}{\shortstack{Gemma~4\\12B}}
 & Append-all & 31.5 & 16.0 \\
 & Mem0 & 60.9 & 52.1 \\
 & EverMemOS-style & 29.7 & 13.2 \\
 & ROAM & \textbf{85.9} & \textbf{79.9} \\
\midrule
\multirow{4}{*}{\shortstack{Qwen3.5\\9B}}
 & Append-all & 31.5 & 16.0 \\
 & Mem0 & 73.3 & 45.1 \\
 & EverMemOS-style & 29.6 & 11.8 \\
 & ROAM & \textbf{83.5} & \textbf{91.0} \\
\bottomrule
\end{tabular}
\end{table}

\subsection{Memory-Management Token Consumption}
\label{sec:appendix-management-tokens}

We measure memory-management token consumption on LongMemEval Controlled at $N=4$ and report averages over 470 episodes. All three methods use the same manager model and a common token-counting convention. The measurements cover management calls for relation inference, memory-operation decisions, and fusion, excluding fact extraction, answer generation, and answer evaluation. Input tokens are counted after prefix-cache reuse; the shared-prefix proportion is measured relative to the input token count before reuse.

\begin{table}[t]
\caption{Mean memory-management overhead per episode at $N=4$. Call counts include failed requests.}
\label{tab:management-token-consumption}
\centering
\small
\setlength{\tabcolsep}{3pt}
\resizebox{\columnwidth}{!}{%
\begin{tabular}{lrrr}
\toprule
\textbf{Metric} & \textbf{Mem0} & \textbf{EverMemOS-style} & \textbf{ROAM} \\
\midrule
LLM calls & 30.1 & 18.6 & 271.6 \\
Input tokens after prefix reuse (k) & 16.79 & 1.56 & 24.79 \\
Output tokens (k) & 11.46 & 0.85 & 1.90 \\
Shared-prefix proportion (\%) & 69.5 & 60.5 & 94.6 \\
\bottomrule
\end{tabular}%
}
\end{table}

ROAM decomposes memory management into fine-grained relation inference, deterministic state updates, and fusion when needed, resulting in more management calls. Nevertheless, it generates 1.90k output tokens per episode, 83.4\% fewer than Mem0, consistent with its use of constrained relation-label outputs. Shared prefixes account for 94.6\% of ROAM's input tokens before reuse, indicating that repeated instructions constitute a substantial portion of its input volume. These results characterize ROAM's design tradeoff: explicit relation management requires more fine-grained calls while keeping the generated output volume small.

\onecolumn
\section{Complete Prompts}
\label{sec:appendix-prompts-config}

The following boxes report the prompts used for baseline memory management,
relation inference, fusion, answer generation, and evaluation. Template
variables are shown in double brackets.

\paragraph{Mem0 direct-operation prompt.}
\begin{promptbox}
You are a smart memory manager which controls the memory of a system.
You can perform four operations: (1) add into the memory, (2) update the memory, (3) delete from the memory, and (4) no change.

Based on the above four operations, the memory will change.

Compare newly retrieved facts with the existing memory. For each new fact, decide whether to:
- ADD: Add it to the memory as a new element
- UPDATE: Update an existing memory element
- DELETE: Delete an existing memory element
- NONE: Make no change (if the fact is already present or irrelevant)

There are specific guidelines to select which operation to perform:

1. **Add**: If the retrieved facts contain new information not present in the memory, then you have to add it by generating a new ID in the id field.
- **Example**:
    - Old Memory:
        [
            {
                "id" : "0",
                "text" : "User is a software engineer"
            }
        ]
    - Retrieved facts: ["Name is John"]
    - New Memory:
        {
            "memory" : [
                {
                    "id" : "0",
                    "text" : "User is a software engineer",
                    "event" : "NONE"
                },
                {
                    "id" : "1",
                    "text" : "Name is John",
                    "event" : "ADD"
                }
            ]

        }

2. **Update**: If the retrieved facts contain information that is already present in the memory but the information is totally different, then you have to update it.
If the retrieved fact contains information that conveys the same thing as the elements present in the memory, then you have to keep the fact which has the most information.
Example (a) -- if the memory contains "User likes to play cricket" and the retrieved fact is "Loves to play cricket with friends", then update the memory with the retrieved facts.
Example (b) -- if the memory contains "Likes cheese pizza" and the retrieved fact is "Loves cheese pizza", then you do not need to update it because they convey the same information.
If the direction is to update the memory, then you have to update it.
Please keep in mind while updating you have to keep the same ID.
Please note to return the IDs in the output from the input IDs only and do not generate any new ID.
- **Example**:
    - Old Memory:
        [
            {
                "id" : "0",
                "text" : "I really like cheese pizza"
            },
            {
                "id" : "1",
                "text" : "User is a software engineer"
            },
            {
                "id" : "2",
                "text" : "User likes to play cricket"
            }
        ]
    - Retrieved facts: ["Loves chicken pizza", "Loves to play cricket with friends"]
    - New Memory:
        {
        "memory" : [
                {
                    "id" : "0",
                    "text" : "Loves cheese and chicken pizza",
                    "event" : "UPDATE",
                    "old_memory" : "I really like cheese pizza"
                },
                {
                    "id" : "1",
                    "text" : "User is a software engineer",
                    "event" : "NONE"
                },
                {
                    "id" : "2",
                    "text" : "Loves to play cricket with friends",
                    "event" : "UPDATE",
                    "old_memory" : "User likes to play cricket"
                }
            ]
        }

3. **Delete**: If the retrieved facts contain information that contradicts the information present in the memory, then you have to delete it. Or if the direction is to delete the memory, then you have to delete it.
Please note to return the IDs in the output from the input IDs only and do not generate any new ID.
- **Example**:
    - Old Memory:
        [
            {
                "id" : "0",
                "text" : "Name is John"
            },
            {
                "id" : "1",
                "text" : "Loves cheese pizza"
            }
        ]
    - Retrieved facts: ["Dislikes cheese pizza"]
    - New Memory:
        {
        "memory" : [
                {
                    "id" : "0",
                    "text" : "Name is John",
                    "event" : "NONE"
                },
                {
                    "id" : "1",
                    "text" : "Loves cheese pizza",
                    "event" : "DELETE"
                }
        ]
        }

4. **No Change**: If the retrieved facts contain information that is already present in the memory, then you do not need to make any changes.
- **Example**:
    - Old Memory:
        [
            {
                "id" : "0",
                "text" : "Name is John"
            },
            {
                "id" : "1",
                "text" : "Loves cheese pizza"
            }
        ]
    - Retrieved facts: ["Name is John"]
    - New Memory:
        {
        "memory" : [
                {
                    "id" : "0",
                    "text" : "Name is John",
                    "event" : "NONE"
                },
                {
                    "id" : "1",
                    "text" : "Loves cheese pizza",
                    "event" : "NONE"
                }
            ]
        }

[[ current_memory_part ]]

The new retrieved facts are mentioned in the triple backticks. You have to analyze the new retrieved facts and determine whether these facts should be added, updated, or deleted in the memory.

```
[[ response_content ]]
```

You must return your response in the following JSON structure only:

{
    "memory" : [
        {
            "id" : "<ID of the memory>",
            "text" : "<Content of the memory>",
            "event" : "<Operation to be performed>",
            "old_memory" : "<Old memory content>"
        },
        ...
    ]
}

Follow the instruction mentioned below:
- Do not return anything from the custom few shot prompts provided above.
- If the current memory is empty, then you have to add the new retrieved facts to the memory.
- You should return the updated memory in only JSON format as shown below. The memory key should be the same if no changes are made.
- If there is an addition, generate a new key and add the new memory corresponding to it.
- If there is a deletion, the memory key-value pair should be removed from the memory.
- If there is an update, the ID key should remain the same and only the value needs to be updated.

Do not return anything except the JSON format.
\end{promptbox}

\paragraph{Prompt for relation labeling and ROAM's relation inference.}
The same prompt is used both to obtain the initial relation labels and to perform relation inference within ROAM.
\begin{promptbox}
# Task

Compare two facts about the SAME user.

- `OLD FACT`: an existing memory.
- `NEW FACT`: a newly extracted candidate fact.

## Label Set (Return Exactly One)

- `IND`: the facts can both be true, are only loosely related, or have no entailment relation.
- `EQV`: the two facts are semantically equivalent (same truth conditions, mutual entailment).
- `OSN`: `NEW FACT` strictly entails `OLD FACT`, so NEW is stronger, narrower, or more specific.
- `NSO`: `OLD FACT` strictly entails `NEW FACT`, so NEW is weaker, broader, or less specific.
- `CON`: the two facts cannot both be true -- incompatible values for the same single-valued attribute of the same aligned entity.

## Decision Process

1. Identify the aligned entity and attribute in each fact. If they are about clearly different entities or different attributes, choose `IND`.
2. If they share the same aligned entity and attribute with **incompatible values**, choose `CON`.
3. If they entail each other (same truth conditions, reworded), choose `EQV`.
4. If only `NEW FACT` entails `OLD FACT` (strict entailment -- NEW being true forces OLD to be true), choose `OSN`.
5. If only `OLD FACT` entails `NEW FACT` (strict entailment -- OLD being true forces NEW to be true), choose `NSO`.
6. Otherwise, choose `IND`.

## What Counts as Each Label

### EQV
Same truth conditions -- each entails the other. They say the same thing, possibly reworded.
- Example: OLD "the user works at Google." / NEW "the user is employed by Google." -> `EQV`.

What does NOT count as `EQV`:
- One fact is strictly more specific than the other (that is `OSN`/`NSO`, not `EQV`).
- The facts conflict (`CON`) -> reject EQV.
- The facts are merely related or about the same topic but not truth-conditionally identical -> reject EQV.
- Different aligned entity or attribute -> not EQV.

### OSN (new entails old)
`NEW FACT` strictly entails `OLD FACT`: if NEW is true, OLD must also be true. NEW adds specificity to OLD (same aligned entity and attribute).
- Example: OLD "the user has a pet." / NEW "the user has a golden retriever named Max." -> `OSN`.
- Example: OLD "the user has a daughter." / NEW "the user has a daughter named Emma." -> `OSN`.
- Geographic containment: OLD "the user lives in New York City." / NEW "the user lives in Manhattan." -> `OSN` (Manhattan is within NYC).

What does NOT count as `OSN`:
- The entailment runs the other way (OLD entails NEW) -- that is `NSO`.
- The two are equivalent (`EQV`) -- not `OSN`.
- Only loose association or plausible implication, not strict entailment -> `IND`.
- Different aligned entity or attribute -> `IND`.
- An added detail that does not actually entail the old fact (e.g., adding an unrelated property) -> `IND`.

### NSO (old entails new)
`OLD FACT` strictly entails `NEW FACT`: if OLD is true, NEW must also be true. NEW is weaker/broader (same aligned entity and attribute).
- Example: OLD "the user lives in Manhattan." / NEW "the user lives in New York City." -> `NSO` (Manhattan is within NYC).
- Example: OLD "the user has a golden retriever." / NEW "the user has a dog." -> `NSO`.
- Geographic containment and clear taxonomic "is-a" relations are acceptable as entailment.

What does NOT count as `NSO`:
- The entailment runs the other way (NEW entails OLD) -- that is `OSN`.
- The two are equivalent (`EQV`) -- not `NSO`.
- Only loose association, not strict entailment: OLD "the user studies computer science." / NEW "the user is a software engineer." -> `IND`.
- Working-in vs living-in, liking vs doing, and similar non-entailing pairs -> `IND`.
- Different aligned entity or attribute -> `IND`.

### CON (contradiction)
Incompatible values for the same **single-valued attribute** of the same aligned entity. Single-valued attributes are those a person normally has only one of: residence, employer, job title, marital status, favorite X, current phone/email, age, etc.

**Time-explainable conflicts still count as `CON`.** Even if the two facts could be reconciled by a change over time (moving, switching jobs, changing preferences), treat them as `CON` -- the newer fact supersedes the older one.
- Example: OLD "the user lives in Beijing." / NEW "the user lives in Shanghai." -> `CON`.
- Example: OLD "the user works at Google." / NEW "the user works at Meta." -> `CON`.
- Example: OLD "the user's favorite color is blue." / NEW "the user's favorite color is red." -> `CON`.

What does NOT count as `CON`:
- Different entities: OLD "the user's brother lives in Beijing." / NEW "the user lives in Shanghai." -> `IND`.
- Multi-valued facts where both can hold: OLD "the user has a son named Max." / NEW "the user has a son named Leo." -> `IND` (could have two sons, no stated uniqueness).
- Facts that are simply independent with no shared single-valued attribute -> `IND`.
- One fact refines or generalizes the other without conflicting (that is `OSN`/`NSO`/`EQV`) -> not `CON`.

### IND (independent)
All other cases: both facts can hold at once, no strict entailment relation, or no shared aligned entity/attribute.
- Different entities: OLD "the user's brother lives in Boston." / NEW "the user lives in Boston." -> `IND`.
- Same entity, no entailment: OLD "the user likes jazz." / NEW "the user has a sister named Sue." -> `IND`.
- Related but non-entailing: OLD "the user studies computer science." / NEW "the user is a software engineer." -> `IND`.
- Preference vs behavior: OLD "the user likes Italian food." / NEW "the user often eats pizza." -> `IND`.
- Different time/context avoiding contradiction: OLD "the user lived in Paris in 2020." / NEW "the user lives in Berlin now." -> `IND`.

## Important Constraints

- Use `OSN` and `NSO` **only for strict entailment**, not weak association or plausible implication. When in doubt, choose `IND`.
- The two facts must share the same aligned entity and attribute for `CON`, `OSN`, `NSO`, or `EQV` to apply. If not, choose `IND`.
- For `CON`: the attribute must be single-valued. Multi-valued/accumulable facts (having multiple children, knowing multiple languages) cannot contradict.
- Time-explainable changes (moves, job changes, preference shifts) are `CON`, not `IND`.

## Output Format

- Return a JSON object with exactly one key: `"relation"`.
- The value must be one of: `IND`, `EQV`, `NSO`, `OSN`, `CON`.
- Do not output extra keys, explanation, or prose.
\end{promptbox}

\paragraph{Fusion prompts.}

\begin{promptbox}
Merge the following into **one memory line**. **Minimal length. Zero information loss.**

- `CURRENT MEMORY`: the accumulated answer memory so far (may already combine several facts).
- `NEW FACT`: a newly confirmed fact that the classifier labeled **equivalent** (`EQV`) to a fact already in the current memory (same truth conditions, just reworded).

Two equivalent statements can mean two different things. Decide which case applies:

1. **Same fact restated** -- a stable attribute or a single past event mentioned again (e.g. "works at Google", "visited Rome in 2019"). State it **once**; do not duplicate phrasings and do not invent a count.
2. **A repeatable event that recurred** -- an action that can naturally happen more than once (e.g. buying snacks, visiting a friend, going to the gym). If `CURRENT MEMORY` and `NEW FACT` describe such an event happening on **different occasions**, record it as **multiple occurrences** with an exact count (e.g. "the user bought spicy snacks twice").

Use the occurrence times to decide:
- `CURRENT MEMORY` time: [[ current_memory_time if current_memory_time else "unknown" ]]
- `NEW FACT` time: [[ new_fact_time if new_fact_time else "unknown" ]]

## Rules

1. **Value chains MUST be preserved.** If `CURRENT MEMORY` contains an arrow-separated history like `v1 -> v2 -> v3 (current)`, copy the ENTIRE chain verbatim.

2. **Counts and numbers are always distinct information.** Never drop a number. Never drop a count.

3. For stable attributes (employment, age, location, preferences, relationships), never count -- always state once. Keep the version that has more specific detail. If both have different but not conflicting details (e.g., one mentions the company, the other mentions the role), merge both details.

4. **Same-attribute check with different values:** If `CURRENT MEMORY` and `NEW FACT` describe the same attribute but with different specific values (even though they are labeled EQV), preserve both: `previously <old_value>; now <new_value>`. This is a safety net -- the classifier should have used CON, but if EQV was chosen, do not lose information.

5. Count repeatable events conservatively:
   - Count as a separate occurrence **only** when the event is genuinely repeatable AND the two times are clearly different occasions.
   - If the times are the same, missing, or it is plausibly the **same** event re-mentioned, do **not** increase the count.
   - When `CURRENT MEMORY` already states a count and `NEW FACT` is the same repeatable event at a **different** time, increment the count directly ("twice" -> "three times").

6. Keep any other unrelated factual content from `CURRENT MEMORY` intact. Do not invent facts.

## Examples

CURRENT: "the user bought spicy snacks" (time 2024-01-01)
NEW: "the user bought spicy snacks" (time 2024-01-08)
-> "the user bought spicy snacks twice"

CURRENT: "the user works at Google" (time 2024-01-01)
NEW: "the user is employed by Google" (time 2024-03-01)
-> "the user works at Google"

CURRENT: "The user's 5K PB: 27:45 -> 26:30 (current, as of 2023/07/30)"
NEW: "the user achieved a personal best 5K time of 26:30" (time 2023/07/30)
-> "The user's 5K PB: 27:45 -> 26:30 (current, as of 2023/07/30)"

CURRENT MEMORY:
[[ current_memory ]]

NEW FACT:
[[ new_fact ]]

Merged line:
\end{promptbox}

\paragraph{EverMemOS-style factual-memory management prompt.}
\begin{promptbox}
You are a memory consolidation assistant. The following facts were extracted from the same conversation and are semantically related -- they belong to the same topic or subject.

Merge them into **one concise, comprehensive sentence or short paragraph** that captures all the information without repetition.

Rules:
- Preserve every distinct piece of information.
- Do not invent facts not present in the list.
- Do not include bullets, headings, or preamble -- output only the merged memory text.
- If the facts are already covered by a single fact, return that fact as-is.

Facts to merge:
[[ facts_text ]]

Merged memory:
\end{promptbox}

\paragraph{Answer-generation prompt.}
\begin{promptbox}
You are a memory-augmented assistant. Use the retrieved memory units to provide accurate and context-aware answers to the user's questions.

[[ context_block ]]

### Question Details
{
- Current Date: [[ question_time ]]
{
- Question: [[ question ]]

Please give a short answer.
\end{promptbox}

\enlargethispage{2\baselineskip}
\paragraph{DeepSeek~V4~Flash answer-judging prompt.}
\begin{promptbox}
You are given a question, its ground-truth answer, and a model response. Judge if the model response is semantically correct. Be lenient for wording differences if the core meaning is correct.

**Question**: [[ question ]]

**Ground-truth answer**: [[ reference ]]

**Model response**: [[ candidate ]]
\end{promptbox}

%% file: sections/02_related_work.tex
\section{Related Work}

\noindent\textbf{Memory representations.}
Long-term memory systems must first decide what unit of past interaction to store and retrieve. Early systems store conversation histories as text and retrieve relevant passages~\citep{packer2023memgpt,zhong2023memorybank}. Yet vector retrieval over raw histories does not itself organize heterogeneous knowledge or determine how that knowledge should evolve~\citep{hatalis2024memory}. A complementary line therefore converts durable information into compact facts or relations. RET-LLM~\citep{modarressi2023retllm} maintains a read--write memory of triples, Dense X Retrieval~\citep{chen-etal-2024-dense} indexes atomic propositions, and HippoRAG~\citep{gutierrez2024hipporag} combines extracted relations with graph-based retrieval. Recent agent-memory systems likewise use atomic facts as retrieval units or enrich them with contextual structure~\citep{shen2026anchormem,huo2026atommem}. Across these systems, fine-grained units provide precise targets for retrieval, comparison, and updating. The same granularity, however, allows redundant, overlapping, and conflicting observations to accumulate as the store grows. ROAM assumes atomic facts have already been extracted and addresses how such observations should coexist, rather than proposing a new extractor, embedding model, or retriever.

\noindent\textbf{Memory management.}
Once information is stored at fine granularity, a second problem is how to maintain the growing collection. One family delegates storage decisions directly to a language model. Mem0~\citep{chhikara2025mem0} prompts the model to add, update, delete, or retain memories, while Memory-R1~\citep{yan2026memoryr1} and AtomMem~\citep{huo2026atommem} learn policies over such operations. Another family emphasizes structural organization or consolidation. A-MEM~\citep{xu2025amem} links memories and augments them with contextual attributes; Reflective Memory Management (RMM)~\citep{tan-etal-2025-prospect} adaptively summarizes dialogue histories and refines retrieval; MemoryOS~\citep{kang-etal-2025-memory} coordinates hierarchical storage, updating, and retrieval; and LightMem~\citep{fang2026lightmem} separates lightweight online processing from offline consolidation. EverMemOS~\citep{hu2026evermemos} further consolidates episodic MemCells into higher-level MemScenes. These approaches position memory management between extraction and retrieval, but expose different control interfaces. Direct-operation policies couple semantic interpretation with storage mutation, whereas structural organizers generally do not derive an observation's retrieval eligibility from an explicit semantic relation. ROAM instead uses that relation as the interface between model judgment and storage control. For each incoming--stored pair, it distinguishes independence, equivalence, conflict, and the two directions of subsumption; a fixed policy then assigns active Primary and supporting Evidence roles. Source observations remain unchanged, while a separate fusion stage constructs compact views for answer-time retrieval.

\noindent\textbf{Evolving memory and evaluation.}
Organization becomes harder when facts change over time, because a memory system must identify the current state without losing earlier evidence. THEANINE~\citep{ong-etal-2025-towards} retains old memories in temporal and causal timelines, Zep~\citep{rasmussen2025zep} represents changing information in a temporal knowledge graph, and APEX-MEM~\citep{banerjee-etal-2026-apex} supports temporal reasoning over semi-structured conversational memory. Supersede~\citep{patel2026supersede} diagnoses failures when agents replace outdated values under a bounded memory budget. Related work also controls which observations enter the store and how stored entries are updated~\citep{zhang2026amac,lam2026ssgm}. Together, these studies motivate temporal structure and explicit update control. ROAM handles temporal conflicts in the same relation space as restatements and specificity refinements, rather than invoking a separate update mechanism. Earlier observations remain available as Evidence, but only views associated with active Primary records are retrieved for answering, separating historical provenance from retrieval eligibility. Evaluation has likewise moved beyond simple recall toward acquisition, updating, and temporal reasoning. LongMemEval~\citep{wu2024longmemeval}, MEME~\citep{jung2026meme}, and MemoryAgentBench~\citep{hu2026memoryagentbench} measure complementary aspects of these capabilities. We use LongMemEval and the post-change portion of MEME, and add a controlled intervention that holds answer-critical memories fixed while varying semantic and temporal competition under a finite retrieval budget.